\documentclass[lettersize,journal]{IEEEtran}
\usepackage{amsmath,amsfonts}
\usepackage{algorithmic}
\usepackage{algorithm}
\usepackage{array}
\usepackage[caption=false,font=normalsize,labelfont=sf,textfont=sf]{subfig}
\usepackage{textcomp}
\usepackage{stfloats}
\usepackage{url}
\usepackage{verbatim}
\usepackage{graphicx}
\usepackage{cite}
\usepackage{booktabs}
\usepackage[utf8]{inputenc}
\usepackage{xcolor}
\usepackage{bbding}
\usepackage{pifont}
\usepackage{wasysym}
\usepackage{utfsym}
\usepackage{fontawesome}
\usepackage{multirow}
\usepackage{colortbl}
\usepackage[colorlinks, linkcolor=blue, anchorcolor=blue, citecolor=blue]{hyperref}
\usepackage{soul, color}

\definecolor{lightgray}{gray}{0.90} 

\definecolor{mycolor}{RGB}{0, 0, 0} 

\definecolor{newcolor}{RGB}{0, 0, 0} 

\usepackage[colorlinks,
            linkcolor=blue,
            anchorcolor=blue,
            citecolor=blue
            ]{hyperref}

\begin{document}

\title{Adaptive Attention Distillation for Robust Few-Shot Segmentation under Environmental Perturbations}

\author{Qianyu Guo, Jingrong Wu, Jieji Ren, Weifeng Ge, Wenqiang Zhang
\thanks{This work was supported by National Natural Science Foundation of China (No.6250070263 and No.52505029), Shanghai Science and Technology Committee (No.25ZR1402293 and No.25ZR1401191 ) , and Shanghai Jiao Tong University (No.YG2025QNB02).
\textit{(Corresponding Authors: Jieji Ren, and Weifeng Ge)}}

\thanks{Qianyu Guo is with the Shanghai Institute of Virology, Shanghai Jiao Tong University School of Medicine, 200025, China. (e-mail: qyguo@sjtu.edu.cn)}

\thanks{Jingrong Wu is with the School of Computer Science and Engineering, Southeast University, Nanjing, 210096, China. (e-mail: candicewu0211@gmail.com)}

\thanks{Jieji Ren is with the School of Mechanical Engineering, Shanghai Jiao Tong University, Shanghai, 200240, China. (e-mail: jiejiren@sjtu.edu.cn)}

\thanks{Weifeng Ge and Wenqiang Zhang are with the Shanghai Key Lab of Intelligent Information Processing, School of Computer Science, Fudan University. Wenqiang Zhang is also with Engineering Research Center of AI \& Robotics, Ministry of Education, Academy for Engineering \& Technology, Fudan University, Shanghai, 200043, China. (e-mail: wfge@fudan.edu.cn and wqzhang@fudan.edu.cn).}
}

\maketitle

\begin{abstract}

Few-shot segmentation (FSS) aims to rapidly learn novel class concepts from limited examples to segment specific targets in unseen images, and has been widely applied in areas such as medical diagnosis and industrial inspection. However, existing studies largely overlook the complex environmental factors encountered in real-world scenarios—such as illumination, background, and camera viewpoint—which can substantially increase the difficulty of test images. As a result, models trained under laboratory conditions often fall short of practical deployment requirements. To bridge this gap, in this paper, an environment-robust FSS setting is introduced that explicitly incorporates challenging test cases arising from complex environments—such as motion blur, small objects, and camouflaged targets—to enhance model's robustness under realistic, dynamic conditions. An environment-robust FSS benchmark (ER-FSS) is established, covering eight datasets across multiple real-world scenarios. In addition, an Adaptive Attention Distillation (AAD) method is proposed, which repeatedly contrasts and distills key shared semantics between known (support) and unknown (query) images to derive class-specific attention for novel categories. This strengthens the model’s ability to focus on the correct targets in complex environments, thereby improving environmental robustness. Comparative experiments show that AAD improves mIoU by 3.3\%–8.5\% across all datasets and settings, demonstrating superior performance and strong generalization. The source code and dataset are available at: https://github.com/guoqianyu-alberta/Adaptive-Attention-Distillation-for-FSS.

\end{abstract}

\begin{IEEEkeywords}
Few-shot segmentation, Environment-robust, Adaptive attention distillation, Benchmark dataset.
\end{IEEEkeywords}

\section{Introduction}
\label{sec:intro}

\begin{figure}[thb]
	\centering
	\includegraphics[width=1.0\linewidth]{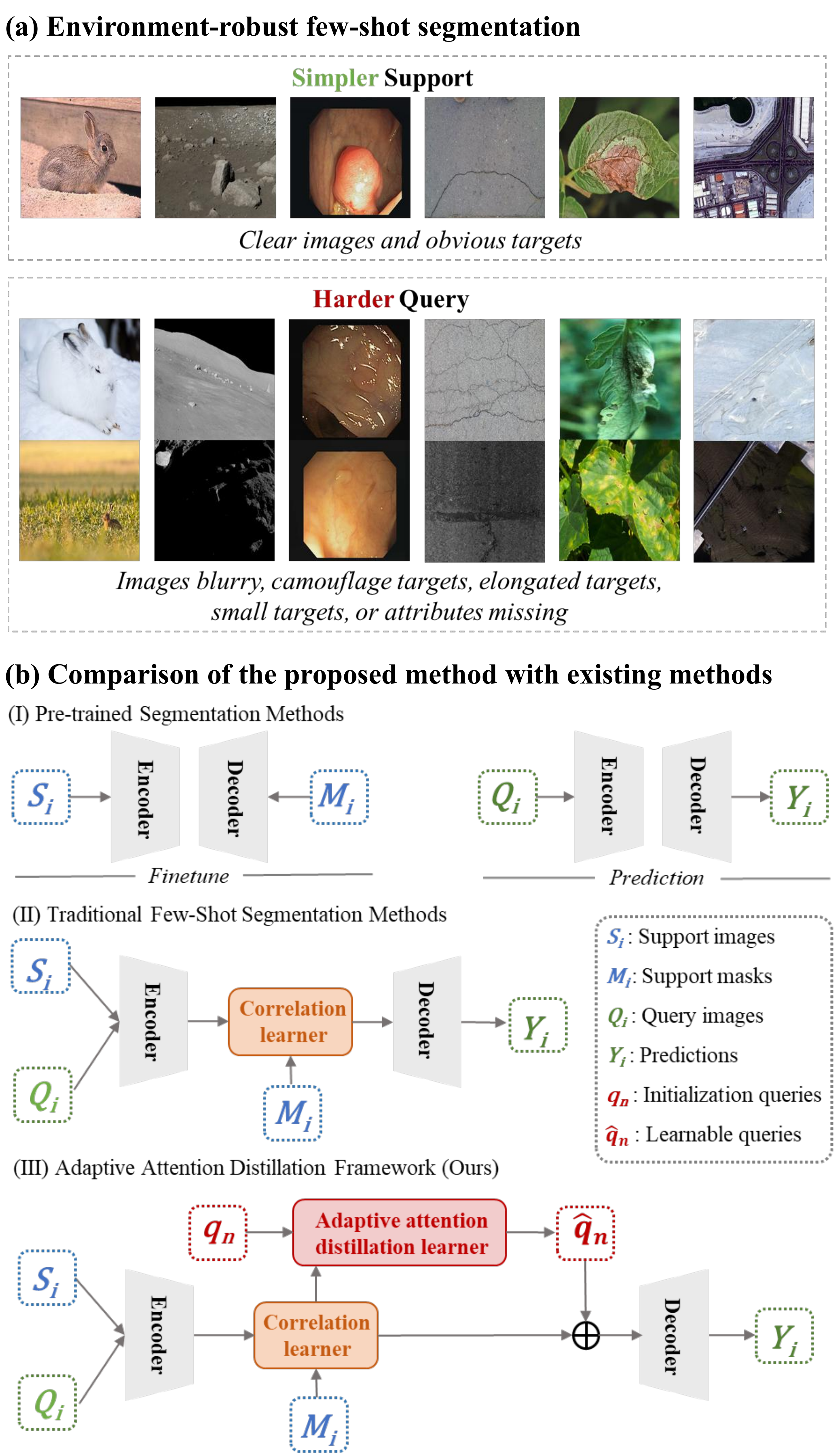}
	\caption{{\color{newcolor}The environment-robust few-shot segmentation (ER-FSS) problem and the solution proposed in this paper: (a) a comparison of environmental difficulty between query and support images in Environment-Robust Few-Shot Segmentation; (b) a comparison between the proposed Adaptive Attention Distillation (AAD) method (\uppercase\expandafter{\romannumeral1}) and existing approaches (pre-training segmentation methods (\uppercase\expandafter{\romannumeral1}) and traditional few-shot segmentation (FSS) methods (\uppercase\expandafter{\romannumeral2})).}}
	\label{Fig:introduction}
\end{figure}

\IEEEPARstart{I}{mage} segmentation serves as a cornerstone in computer vision, underpinning critical applications ranging from medical diagnosis to aerospace analysis~\cite{abs-2308-16184, CaiFZF25, WuZPYSTL25, abs-2306-16269, GengSJ24}. However, the prohibitive cost of acquiring pixel-level annotations for large-scale datasets has catalyzed research into few-shot segmentation (FSS), which aims to segment novel categories given only a handful of labeled examples. This data-efficient paradigm has proven particularly valuable in specialized domains where labeled data is scarce. For instance, in medical imaging, recent methods~\cite{DissanayakeGMSFG26, SilvaRodriguezDA25} leverage anatomical priors or volumetric consistency to handle diverse modalities; in remote sensing, approaches~\cite{YangTMZPJ24, YangXXTWSY25} have been tailored to address significant scale variations and complex geospatial contexts; similarly, in industrial inspection, FSS~\cite{ZhangNZZC25} are increasingly adopted for detecting surface defects under limited sample conditions.

A prevalent strategy for label-efficient segmentation is the pretrain--finetune paradigm, wherein models pre-trained on large-scale data (e.g., Swin Transformer~\cite{LiuL00W0LG21}, SAM~\cite{abs-2304-02643}) are adapted to downstream tasks. However, in regimes of severe data scarcity, fine-tuning is prone to overfitting. FSS circumvents this limitation by learning transferable patterns from a limited set of support images to segment unseen query instances. The majority of FSS approaches leverage a meta-learning framework underpinned by Siamese or prototypical architectures~\cite{ShiWZLNCMZ22, WangLZZF19, MinKC21, ZhangKYW21, LiuJY21, BoudiafKZPAD21, ChengLH23, GaoFHWLY22}. Recent advancements~\cite{abs-2308-09294, ShiWZLNCMZ22} have demonstrated the efficacy of these methods, achieving nearly $70\%$ mIoU in the 1-shot setting on standard benchmarks such as PASCAL-5$^i$~\cite{ShabanBLEB17} and COCO-20$^i$~\cite{NguyenT19}.

Despite this progress, real-world conditions introduce complex environmental variations—such as illumination changes, cluttered backgrounds, object motion, and viewpoint shifts—that significantly increase the difficulty of query images compared to support images. These factors can obscure target boundaries, distort shapes, or cause severe blur, resulting in a sharp degradation of FSS performance outside controlled environments. Unfortunately, most existing studies, datasets, and models overlook these real-world challenges, ultimately limiting the practical deployment of FSS algorithms.

{\color{newcolor}To address these challenges, the Environment-robust Few-shot Segmentation (ER-FSS) task (see Fig.~\ref{Fig:introduction}(a)) is introduced to enhance the resilience of FSS models against environmental perturbations. This task specifically targets typical hard cases in query images arising from complex real-world conditions, such as motion blur, small objects, camouflaged targets, and the occlusion of key features. To better simulate practical application scenarios, images exhibiting these challenges are employed as queries, whereas simpler, cleaner samples captured under controlled settings serve as support images. Built upon this paradigm, the ER-FSS benchmark is constructed, encompassing six scenario types and eight datasets. Unlike conventional datasets, ER-FSS provides a more faithful reflection of model performance, generalization, and robustness across diverse environmental variations and domains, thereby offering a highly realistic benchmark for evaluating both pre-trained segmentation and FSS models.}

Evaluations on the ER-FSS benchmark reveal that state-of-the-art (SOTA) pre-training--fine-tuning and FSS models exhibit limited robustness to environmental variations, falling significantly short of practical deployment requirements. 
{\color{newcolor}To mitigate this issue, Adaptive Attention Distillation (AAD) (see Fig.~\ref{Fig:introduction}(b)) is proposed. 
It acts as a progressive feature correction process by iteratively contrasting semantic information between the support and query images. Across multiple refinement stages, learnable queries extract and distill the purest, class-discriminative semantics while actively discarding background-induced variations. The resulting distilled attention serves as a potent corrective map; by re-weighting the feature maps, it heavily suppresses background activations and effectively pulls the deviated feature representations back to the correct target manifold. By consolidating these critical semantic cues and calibrating the features, AAD significantly enhances target localization and segmentation accuracy in complex environments.}
Experimental results demonstrate that AAD consistently outperforms existing pre-trained and FSS models, yielding an average Intersection over Union (mIoU) improvement of 3.3\%--8.75\% over current SOTA approaches.

In summary, this work makes the following contributions:
\begin{itemize}
\item Introduces the ER-FSS task and the accompanying benchmark, consisting of eight datasets across diverse scenarios to enable realistic, multi-scene evaluation of segmentation robustness.
\item Proposes Adaptive Attention Distillation (AAD), which iteratively contrasts semantic information between support and query images to distill class-specific attention, thereby improving target recognition and enhancing robustness under challenging environmental conditions.
\item Extensive experiments on ER-FSS show that AAD significantly outperforms existing FSS and pretrain–finetune models, achieving stronger generalization and higher robustness across a wide range of scenarios and settings.
\end{itemize}

\section{Related Work}
\label{sec:related}

\subsection{Few-Shot Segmentation (FSS)}
Standard FSS~\cite{ShabanBLEB17, NguyenT19, LiWCTT20, MinKC21, BoudiafKZPAD21, ChengLH23, TianZSYLJ22, ZhangKYW21, GaoZJLW25, LiuLWCAYH25, LuoCCIK25} typically follows a meta-learning paradigm involving two distinct phases: training on source-domain base classes and inference on target-domain novel classes. During the training phase, the model learns generalizable feature representations from abundant annotated data of base classes. In the inference phase, the goal is to segment query images containing unseen categories, guided by only a few annotated support samples from the target domain.

OSLSM~\cite{ShabanBLEB17} pioneered this field by dynamically generating classifier weights for each query-support pair. Inspired by Prototypical Networks~\cite{SnellSZ17}, most contemporary FSS methods adopt a dual-branch meta-learning framework, broadly categorized into prototypical feature learning and relation-based methods. The former, exemplified by PANet~\cite{WangLZZF19}, focuses on optimizing prototype representations to better separate foreground from background, while the latter emphasizes refining similarity metrics between extracted features.However, existing approaches often overlook a critical practical asymmetry: query images are frequently significantly more complex than support samples. When query targets suffer from severe occlusion or lack distinct category cues, standard models struggle to accurately localize the object. 

\subsection{Related Datasets}
\label{sub:related}
Standard FSS evaluations predominantly rely on PASCAL-5$^i$~\cite{ShabanBLEB17} and COCO-20$^i$~\cite{NguyenT19}, which comprise common object categories. During testing, support and query sets are randomly sampled from novel classes; however, these datasets largely fail to capture the complexity inherent in real-world deployment. To address generalization, the Cross-Domain Few-Shot Segmentation (CD-FSS) benchmark~\cite{LeiZHCDL22} was proposed, incorporating diverse domains such as medical imaging~\cite{CandemirJPMSXKATM14, abs-1902-03368} and agriculture~\cite{DemirKLPHBHTR18}. 

Yet, CD-FSS remains limited in scope. With the exception of Deepglobe, performance on CD-FSS datasets mirrors that of general domains, suggesting it does not impose sufficient difficulty to expose the vulnerabilities of modern models. Crucially, like its predecessors, CD-FSS overlooks severe environmental perturbations encountered in the wild. To bridge this gap, we introduce ER-FSS, a rigorous benchmark spanning eight challenging scenarios (e.g., camouflaged and small targets) to evaluate FSS model robustness under realistic, adversarial conditions.


\section{Benchmark Dataset}
\label{sec:benchmark}

\begin{figure*}[thb]
	\centering
	\includegraphics[width=1.0\linewidth]{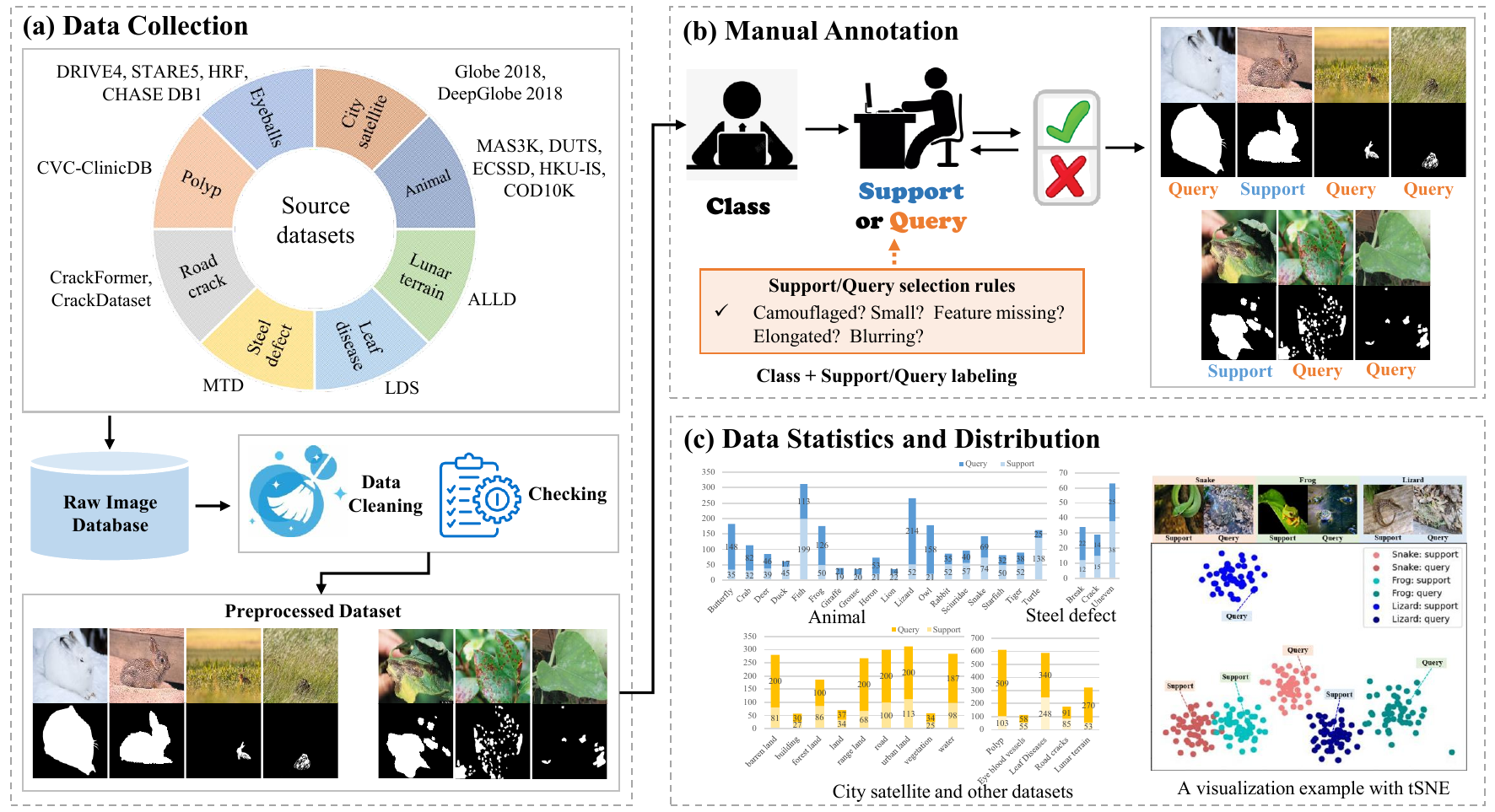}
	\caption{The construction process of  ER-FSS Benchmark: data collection phase (a) and manual annotation phase (b). Data statistics (c) and a visualization example with t-SNE for the evaluation benchmark datasets.}
	\label{Fig:benchmark}
\end{figure*}

\subsection{Overview of the Benchmark Dataset}

In this work, we introduce the ER-FSS benchmark, a comprehensive platform designed to evaluate segmentation algorithms under diverse, realistic conditions. ER-FSS distinguishes itself from prior benchmarks through three key advantages: broad domain coverage, meticulous annotation quality, and a rigorous stratification of environmental complexity.

First, ER-FSS spans six distinct domains across eight datasets, covering: Biology (18 animal categories), Astronomy (lunar terrain features), Medicine (polyps and retinal vessels), Industry (road cracks and steel defects), Agriculture (leaf diseases), and Geography (satellite imagery of urban structures).

Second, to ensure high-quality supervision, every image underwent rigorous manual inspection and cleaning. Third, and most critically, ER-FSS explicitly models the domain gap inherent in real-world applications. We categorize images based on environmental difficulty: clear, laboratory-quality samples are designated as the support set, while challenging, in-the-wild samples serve as the query set. These challenges are formalized into five characteristics: camouflaged objects, small targets, elongated structures, missing attributes, and motion blur. This structured split ensures that models are evaluated on their ability to generalize from simple examples to complex, adversarial scenarios.

\subsection{Construction Process}
As illustrated in Fig.~\ref{Fig:benchmark}, building the ER-FSS benchmark dataset comprises two primary stages: data collection and manual annotation.

\noindent\textbf{Data Collection.} In the data collection phase, we aimed to gather images from as many domains and sources as possible. To this end, we selected images from 17 diverse datasets. The sources for each dataset are as follows: Animals (MAS3K~\cite{LiRDC20}, DUTS~\cite{WangLWF0YR17}, ECSSD~\cite{ShiYXJ16}, IS~\cite{LiY15}, COD10K~\cite{fan2020Camouflage}), Lunar terrain (ALLD), Polyp (CVC-ClinicDB), Eyeballs (DRIVE, STARE, STAREHRF~\cite{BudaiBMHM13}, CHASE DB), Road cracks (CrackForest~\cite{shi2016automatic}, CrackDataset~\cite{AmhazCIB16}), Steel defects (MTD~\cite{8560423}), Leaf diseases (LDS), and City satellite (DeepGlobe 2018~\cite{DemirKLPHBHTR18}, AIS). Following data collection, we performed rigorous cleaning and verification of the images. We ensured that each image had an accurate and valid corresponding mask label, removing any samples with incomplete or incorrect annotations.

\noindent\textbf{Manual Annotation.}
We focus on two primary aspects in our annotation process:
characteristics are defined as follows:
\begin{itemize}
    \item Manual Annotation of Class Labels: For images without category labels, we manually annotate the class texts and rectify errors in any existing labels.
    \item Query/Support Sample Selection: Images exhibiting at least one of the challenging characteristics listed below are categorized as query images, while those that do not are defined as support images.
\end{itemize}

{\color{newcolor}To ensure high-quality annotations, the class label of each image is verified by at least two annotators. The categorization of simple versus difficult samples (i.e., support and query set selection) is reviewed by at least three annotators. In cases of discrepancy, labels are finalized through a majority vote. Because the annotation guidelines are relatively quantified, disagreements during the annotation process are infrequent, thereby ensuring the overall quality and consistency of our dataset. The challenging characteristics are defined as follows:}

\begin{itemize}
    \item Camouflaged Objects: The target and background share similar visual attributes, such as color or texture, making it difficult for both models and humans to distinguish between them.
    \item Small Targets: Targets are considered small if they occupy less than approximately 1$\%$ of the total pixels.
    \item Elongated Targets: Targets with extremely elongated and irregular shapes (e.g., fine retinal blood vessels), which are difficult for models to accurately capture.
    \item Missing Attributes: Crucial distinguishing features of the target are either occluded by other objects or missing due to incomplete capture.
    \item Image Blurring: Reduced image clarity, often due to low resolution or motion blur, which makes target identification challenging.
\end{itemize}

\noindent\textbf{Data Distribution and Statistics.} To ensure robust evaluation, we maintain a minimum of 20 support images and 10 query images per category, with query samples incorporating multiple challenging attributes to enhance diversity. {\color{newcolor}As illustrated in Fig.~\ref{Fig:benchmark}, we detail the specific number of query and support samples for each category across the various datasets. Also, we validate the proposed difficulty stratification via t-SNE visualization of ViT~\cite{DosovitskiyB0WZ21} features.} The analysis reveals a significant distribution shift between easy (support) and hard (query) samples within the same category. For instance, hard samples of ``frog" cluster closer to simple ``lizard" samples than to their own class prototypes. This intra-class divergence and inter-class confusion highlight the critical real-world challenge where models misclassify targets due to feature drift.

Furthermore, ER-FSS captures domain-specific imbalances inherent in practical applications. The challenge distribution varies naturally across datasets: industrial and agricultural subsets (e.g., Road Cracks, Steel Defects, Leaf Diseases) are dominated by slender structures and motion blur; biological data (Animals) emphasizes camouflaged and partially occluded targets; medical imagery (Eyeballs, Polyp) suffers from lighting variations and blur; while remote sensing domains (Lunar Terrain, City Satellite) predominantly feature small-scale objects. This heterogeneity ensures that ER-FSS provides a holistic assessment of model robustness across varying environmental complexities.

\section{Method}

\begin{figure*}[thb]
	\centering
	\includegraphics[width=1.0\linewidth]{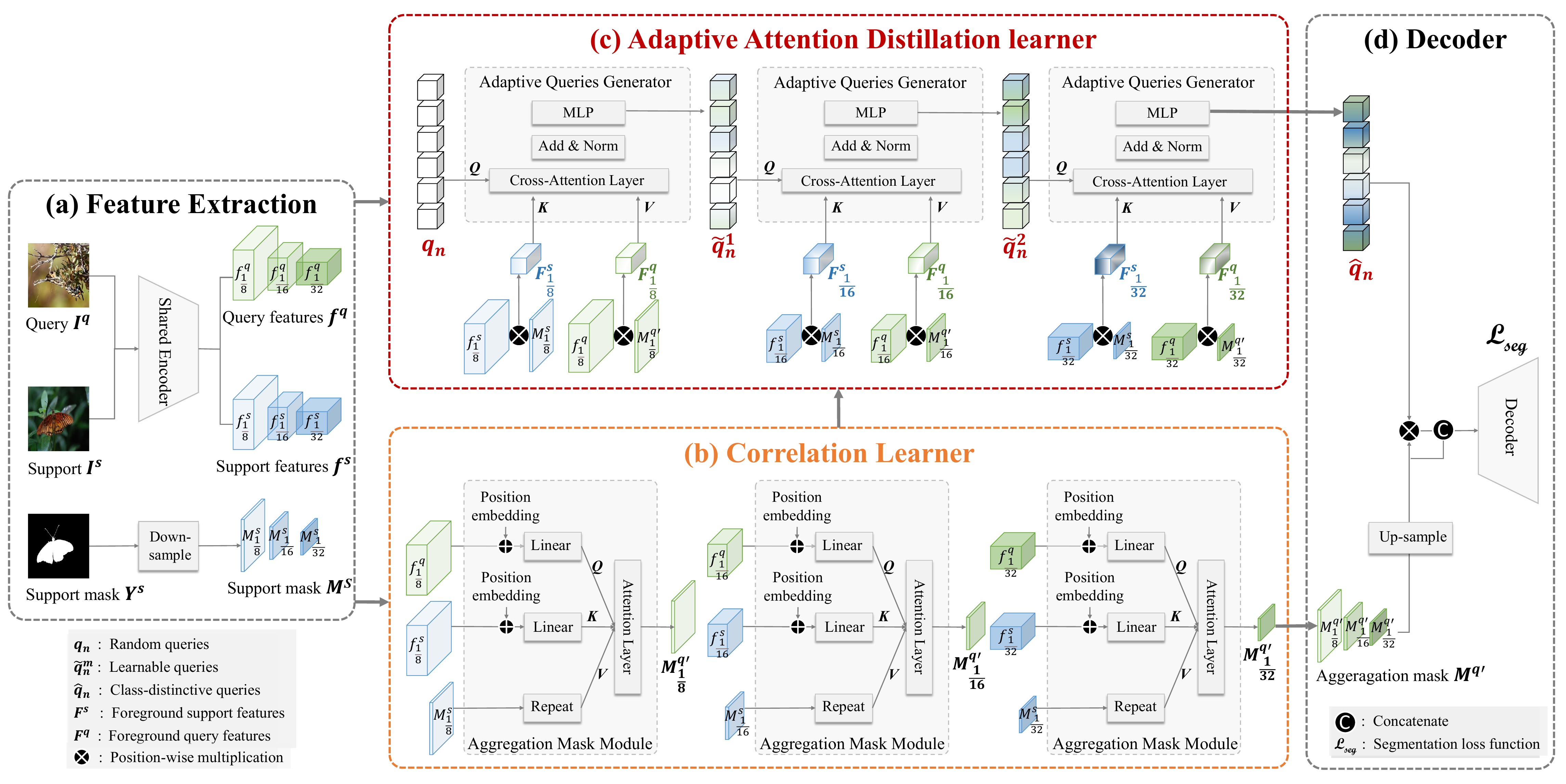}
	\caption{The pipeline of the proposed  Adaptive Attention Distillation (AAD)  framework. AAD framework consists of four parts: the encoder, correlation leaner module,  adaptive attention distillation learner, and the decoder.}
	\label{Fig:method}
\end{figure*}

\subsection{Problem Setting}

We formulate the ER-FSS problem building upon the classic FSS framework. Given a source domain $\mathcal{D}_{\text{train}} = \{(X_P, Y_P)\}$ and a target domain $\mathcal{D}_{\text{test}} = \{(X_T, Y_T)\}$, we assume a distributional shift between domains ($X_P \neq X_T$) and disjoint label spaces ($Y_P \cap Y_T = \emptyset$). The model acquires meta-knowledge on $\mathcal{D}_{\text{train}}$ and is evaluated on $\mathcal{D}_{\text{test}}$ following the episodic paradigm~\cite{VinyalsBLKW16}.

Specifically, for an $N$-way $K$-shot task, each episode comprises a support set $\mathcal{S} = \{(I^s_i, M^s_i)\}_{i=1}^{N \times K}$ and a query set $\mathcal{Q} = \{(I^q_j, M^q_j)\}_{j=1}^{n}$. A critical distinction in our ER-FSS setting is the stratification of difficulty: $\mathcal{S}$ is composed exclusively of simple samples , whereas $\mathcal{Q}$ consists of hard samples. This asymmetric design rigorously evaluates the model's ability to generalize from ideal references to realistic, challenging environments.

Note that in our experiments, we adopt distinct training protocols tailored to each model architecture. For FSS baselines and our method, training strictly follows the episodic meta-learning paradigm. Conversely, for pre-trained models based on transfer learning (e.g., SAM), we adhere to their native single-branch, end-to-end approach. Specifically, these models leverage general representations learned from large-scale datasets (e.g., ImageNet). During the $N$-shot evaluation, we adapt them to the target task by fine-tuning the decoder on the provided $N$ support images, while keeping the backbone encoder frozen to preserve generalization.

\subsection{Method Overview}

{\color{newcolor}Current FSS methods typically struggle when environmental complexity (e.g., camouflage or occlusion, as shown in Fig.~\ref{Fig:benchmark}) introduces massive intra-class variance. This discrepancy inevitably leads to the extraction of biased features, causing models to fail in accurately localizing targets. To mitigate this, we introduce Adaptive Attention Distillation (AAD). Unlike existing paradigms that directly rely on static prototypes or raw cross-attention outputs, our method utilizes initial correlation matching merely as a spatial prior. 

Building upon this, our AAD learner employs learnable queries to iteratively interact with both support and query features. From a representation perspective, this process functions as a progressive feature calibration mechanism. 
It achieves representation alignment by explicitly aligning noisy query features with essential target characteristics, while performing feature smoothing to suppress environment-induced high-frequency noise.
By actively ``distilling'' stable, class-level attention, AAD heavily suppresses background interference and pulls the deviated representations back to the correct target manifold. Consequently, highly calibrated and accurate features are fed into the decoder, empowering the model to maintain strong environmental invariance even in severely degraded scenarios. As depicted in Fig.~\ref{Fig:method}, our overall framework comprises four key modules: a shared encoder, a correlation learner, the AAD learner, and a decoder.}

\subsection{Shared Feature Encoder}

As illustrated in Fig.~\ref{Fig:method}(a), the AAD framework begins with a feature encoder. This module, which can be implemented with a standard backbone architecture such as ResNet, VGG, or a Vision Transformer, adheres to a parameter-sharing paradigm. This design choice is crucial for few-shot learning, as it ensures that both the support image \(I^{s}\) and the query image \(I^{q}\) are projected into a common, consistent feature space.

During each training iteration, a support-query pair \((I^{s}, I^{q})\) is sampled from the training dataset \(D_{\text{train}}\). These images are processed by the shared backbone to produce their respective feature representations, \(f^{s}\) and \(f^{q}\). To facilitate robust matching across different levels of abstraction, we extract multi-scale feature maps. Specifically, for each image, we obtain features \(f_{i}\) at three different scales \(i \in \{\frac{1}{8}, \frac{1}{16}, \frac{1}{32}\}\), where the scale indicates the spatial resolution relative to the original input image size. Thus, \(f^{s}_{i}\) and \(f^{q}_{i}\) denote the feature maps from the \(i\)-th scale, each with a dimension of \(\mathbb{R}^{H_{i} \times W_{i} \times d_{i}}\). Concurrently, the ground-truth segmentation mask of the support image, \(Y^{s}\), is downsampled to match the spatial dimensions of the multi-scale feature maps. This results in a set of support masks, \(M^{s}_{i}\), for each scale. 


\subsection{Correlation Learner}

{\color{newcolor}As shown in Fig.~\ref{Fig:method}(b), the correlation learning (CL) module is designed to establish an initial correspondence between the support and query images. Inspired by the dense cross-attention mechanism introduced in DCAMA~\cite{ShiWZLNCMZ22}, we adopt this concept to capture feature similarities across different inputs. However, to explicitly tailor this mechanism for our proposed Adaptive Attention Distillation (AAD) framework, we make specific architectural adaptations. The module is structured with multiple Aggregation Mask Modules (AMMs), each operating at a specific feature scale. Unlike the original formulation in DCAMA which relies on the attention output for the direct final prediction, the primary function of our AMM is adapted to generate a coarse segmentation mask that serves as a prior. 

The process begins with the AMM at each scale $i$ taking the support features $f^{s}_{i}$ and query features $f^{q}_{i}$ as input. The cross-attention mechanism computes the similarity between them, effectively assigning higher weights to regions in the query features $f^{q}_{i}$ that closely correspond to the target features present in $f^{s}_{i}$. Subsequently, the module incorporates the corresponding support mask $M^{s}_{i}$ to filter and refine these weighted features, producing a coarse mask that approximates the target's location in the query image. These multi-scale coarse segmentation maps serve a dual purpose, which constitutes the core of our structural adaptation. First, they are forwarded to the decoder and progressively upsampled to contribute to the final, high-resolution segmentation result. Second, and most critically for our AAD framework, they provide the essential initial target localization priors required by the subsequent AAD learner. By seamlessly coupling the cross-attention-based CL module with the AAD learner, this architectural adaptation ensures that the AAD learner receives a robust initial understanding of the target, which it then further distills and refines. }

Specifically, for a given scale \(i\), the AMM first reshapes \(f^{s}_{i}\) and \(f^{q}_{i}\) from \(\mathbb{R}^{H_{i}\times W_{i} \times d_{i}}\) to a flattened format of \(\mathbb{R}^{(H_{i}\times W_{i}) \times d_{i}}\) to prepare them for matrix multiplication. The module then computes an attention-guided query mask using scaled dot-product attention, as defined in the following equation:
\begin{equation}\label{eq:amm}
\text{Attention}(f^{q}_{i}, f^{s}_{i}, M^{s}_{i}) = \text{softmax}\left(\frac{f^{q}_{i} (f^{s}_{i})^{T}}{\sqrt{d_{i}}}\right) M^{s}_{i}.
\end{equation}
In this formulation, the query \(f^{q}_{i}\) function as the ``Query" (Q), while the support \(f^{s}_{i}\) serve as the``Key" (K). The dot product \(f^{q}_{i} (f^{s}_{i})^{T}\) calculates a raw similarity matrix between every feature vector in the query and every feature vector in the support. The resulting attention map is then multiplied by the downsampled support mask \(M^{s}_{i}\), which acts as the ``Value" (V). Since \(M^{s}_{i}\) contains binary values indicating the target object's location (1 for the target, 0 for the background), this final matrix multiplication effectively ``filters" the attention scores. It retains and aggregates the attention weights corresponding only to the target category, thereby producing a coarse probability map for the target's location within the query image, which we denote as the coarse query mask \(M^{q'}_{i}\).


\subsection{Adaptive Attention Distillation Learner}

{\color{newcolor}While the CL provides an initial, coarse localization of the target, its reliance on direct feature matching renders it susceptible to failure in hard" cases characterized by significant appearance discrepancies (e.g., camouflage, motion blur, or viewpoint shifts). To transcend these limitations, a mechanism is required to distill essential, class-discriminative semantics independent of low-level variations. The core philosophy behind our Adaptive Attention Distillation (AAD) design is to adaptively learn an attention mechanism focused specifically on the target object, while crucially ensuring this process remains lightweight and rapid to avoid significant penalties in computational load or inference time. Guided by this philosophy, we introduce the AAD learner, illustrated in Fig.~\ref{Fig:method}(c). Unlike dense pixel-to-pixel comparisons, this module innovatively employs lightweight, learnable vectorized queries—functioning as compact, highly informative class queries''—to abstract class-level information and encapsulate the core characteristics of the target category. The module is composed of multiple Adaptive Query Generators (AQGs), which progressively interact with target features from both support and query images. Through this interaction, the AAD learner distills precise attention representations, focusing on fundamental target properties while effectively suppressing distracting background clutter.}

The process commences by isolating foreground features to ensure the learner focuses exclusively on the target object. Using feature maps \(f^{s}_{i}\) and \(f^{q}_{i}\) from the encoder, alongside masks \(M^{s}_{i}\) and \(M^{q'}_{i}\) (derived from the support set and the preceding stage, respectively), we compute the foreground representations \(F^{s}_{i}\) and \(F^{q}_{i}\) via element-wise multiplication:
\begin{equation}\label{eq:foreground}
F^{s}_{i} = f^{s}_{i} \otimes M^{s}_{i}, \quad F^{q}_{i} = f^{q}_{i} \otimes M^{q'}_{i}.
\end{equation}
This operation effectively masks out the background, minimizing its influence on the subsequent distillation process.

{\color{mycolor}Distinct from traditional approaches that rely on static parameters, our query initialization is dynamic. We begin by generating a set of \(N\) random vectors, denoted as \(q_{in} \in \mathbb{R}^{N \times l}\), where \(l\) is the hidden dimension. In our experiments, a small number of queries (e.g., \(N=15\)) proves sufficient. Although these queries originate from a random distribution, they function as adaptive ``information collectors." Through the forward pass, they act as ``learnable" entities in the sense that they adaptively aggregate and refine semantic information from the support and query features, evolving from random noise into discriminative descriptors.}

The core of the AQG is a cross-attention mechanism designed to extract relevant features. Here, the class queries \(q\) serve as the ``Query" (Q), the support foreground features \(F^{s}_{i}\) act as the ``Key" (K), and the query foreground features \(F^{q}_{i}\) serve as the ``Value" (V). This formulation addresses the question: ``Which parts of the query foreground \(F^{q}_{i}\) are most relevant, given the class context provided by the support foreground \(F^{s}_{i}\)?" The updated queries are computed as follows:
\begin{equation}\label{eq:transformer}
\tilde{q}_{i} = \text{MLP}\left(\text{LayerNorm}\left(\text{softmax}\left(\frac{q (F^{s}_{i})^{T}}{\sqrt{d_{i}}}\right)F^{q}_{i} + q\right)\right),
\end{equation}
where the output undergoes layer normalization and projection via an MLP to restore the dimensionality to \(N \times l\). This design allows the queries to abstract the essential what (class identity) rather than the incidental where (pixel location), ensuring robustness against intra-class variations and noise in the coarse mask \(M^{q'}_{i}\).

Crucially, we exploit the hierarchical nature of deep features to progressively refine these queries. The process is iterative across different feature scales. The initially random query set \(q_{in}\) first enters the AAD module at the \(\frac{1}{8}\) scale, interacting with low-level texture and edge details to form the preliminary feature \(q_1\). Subsequently, \(q_1\) propagates to the second and third AAD modules, interacting further with features at the \(\frac{1}{16}\) and \(\frac{1}{32}\) scales, respectively. This multi-stage evolution allows the queries to integrate information ranging from structural details to abstract semantics. Ultimately, the final output \(\hat{q}\) represents a set of highly focused, environment-invariant target attention features, ready to guide the final segmentation decoder.

\subsection{Decoder and Loss Function}

The final stage of the AAD framework employs a decoder to synthesize the coarse localization from the Correlation Learner with the high-level semantic knowledge distilled by the AAD learner. As illustrated in Fig.~\ref{Fig:method}(d), the inputs are the multi-scale coarse query masks \(M^{q'}_{i}\) and the final set of discriminative class queries \(\hat{q}\). To effectively integrate these inputs, we first refine the coarse masks via element-wise multiplication with \(\hat{q}\). This operation re-weights the spatial features, amplifying regions consistent with the distilled class semantics while suppressing background noise. The refined features are then concatenated with the original coarse masks to yield a fused representation \(R_{q}\):
\begin{equation}\label{eq:fusion}
R_{q} = \text{concat}(M^{q'}_{i}, M^{q'}_{i} \otimes \hat{q}).
\end{equation}

{\color{mycolor}This fused representation is processed by a decoder composed of three sequential mixing modules. Each module primarily utilizes two convolutional layers coupled with ReLU activations to progressively compress channel dimensions (from \(256+\text{num\_queries} \to 128 \to 64 \to 16 \to 2\)) while refining spatial details. Through iterative upsampling, the decoder restores the feature maps to the original image resolution to generate the final prediction.} The entire framework is trained end-to-end by minimizing the standard Binary Cross-Entropy (BCE) loss between the predicted segmentation map and the ground-truth query mask.

\section{Experiments}
\label{sec:experiment}

\begin{table*}[th]
	\centering
  	\caption{{\color{mycolor}Comparison with SOTA methods on 1-shot, 5-shot, and 20-shot setting on Road cracks (Industrial), Steel defects (Industrial), and Leaf diseases (Agriculture). The numbers in \textbf{Bold}  indicate the best performance.}}
  \scalebox{1.0}{
	\begin{tabular}{ll|ccc|ccc|ccc}
\hline
\hline
Backbone &Method  &\multicolumn{3}{c|}{Road crack} &\multicolumn{3}{c|}{Steel defect}&\multicolumn{3}{c}{Leaf diseases}\\
&&1-shot &5-shot &20-shot &1-shot &5-shot &20-shot &1-shot &5-shot &20-shot\\
\hline
\multirow{8}{*}{ResNet-50}  
&PFENet{~\color{gray}
{[TPAMI2022]}}~\cite{TianZSYLJ22} & 0.44 &0.41 & 0.39 &9.06 &9.06 &9.14  &11.60 &11.06 &11.13
\\
  &DCAMA{~\color{gray}{[ECCV2022]}}~\cite{ShiWZLNCMZ22} &3.36 &5.23 &6.26 &7.20 &7.13 &11.80 &19.80 &22.00 &23.29\\
&HDMNet{~\color{gray}{[CVPR2023]}}~\cite{PengTWWLSJ23} &1.20 &1.28 &2.34 &6.43 &6.01 &14.29  &17.82 &18.70 &16.18\\

&PFENet++{~\color{gray}{[TAPMI2024]}}~\cite{LuoTZYTJ24} &0.00 &0.91 &0.00 &5.69 &6.98 &7.70  &17.91 &16.91 &18.89\\
&HMNet{~\color{gray}{[NIPS2024]}}~\cite{XuLZL00024}   &1.20	&1.28	&2.34	&6.43	&6.01	&15.36	&20.87	&21.72	&24.12
\\
&ABCDFSS{~\color{gray}{[CVPR2024]}}~\cite{Herzog24} &5.42	&5.95	&8.03	&7.90	&12.14	&12.34	&21.94	&20.65	&26.78
\\
&NTRENet++{~\color{gray}{[TCSVT2025]}}~\cite{LiuLWCAYH25} &0.54	&3.39	&3.32	&4.41	&12.58	&\textbf{32.48}	&6.96	&18.07	&23.47\\
 &\textbf{AAD (Ours)} &\textbf{8.15} &\textbf{10.22} &\textbf{11.67} &\textbf{10.44} &\textbf{16.66} &24.05 &\textbf{24.57} &\textbf{29.03} &\textbf{30.95} \\
 \hline
 \multirow{3}{*}{ResNet-101} 

  &DCAMA{~\color{gray}{[ECCV2022]}}~\cite{ShiWZLNCMZ22} &1.55 &1.60 &1.62 &8.08 &9.06 &17.52  &22.91 &26.47 &28.23\\
 &SCCAN{~\color{gray}{[ICCV2023]}}\cite{abs-2308-09294} &0.48 &2.27 &1.12 &9.72 &18.16 &14.21 &19.23 &18.23 &16.05\\
 &\textbf{AAD (Ours)}  &\textbf{8.41} &\textbf{10.65} &\textbf{12.18} &\textbf{15.36} &\textbf{24.68} &\textbf{27.97} &\textbf{25.13} &\textbf{30.31} &\textbf{32.31}\\
  \hline
   \multirow{8}{*}{SWin-T-base} 
&SAM{~\color{gray}{[arxiv2023]}}~\cite{abs-2304-02643}  &1.02 &1.02 &1.02 &5.55 &5.55 &14.84 &15.51&15.51&15.51\\
  &DCAMA{~\color{gray}{[ECCV2022]}}~\cite{ShiWZLNCMZ22}&\textbf{11.67} &11.62&12.23 &12.25 &15.28 &30.42 &27.73 &29.44 &30.46\\
  &Matcher{~\color{gray}{[ICLR2024]}}~\cite{LiuZL0WS24}&1.62&1.89 &--&7.76 &8.17 &-- &15.23 &15.26 &--\\
  &MGF-SAM{~\color{gray}{[NIPS2024]}}~\cite{ZhangGJLW24}&3.66	&5.05&4.84 &8.99&10.32	&22.07 &15.81	&21.09	&23.54\\
&GRPN{~\color{gray}{[AAAI2025]}}~\cite{peng2025sam}&10.48&10.37&10.64 &12.50	&18.18	&30.85 &20.44	&27.37	&29.10\\
&FCP{~\color{gray}{[AAAI2025]}}~\cite{park2025foreground}&2.77	&2.99	&3.08&9.14	&13.62	&14.61&17.73	&17.85	&17.81\\
 &FS-SAM{~\color{gray}{[ICML2025]}}~\cite{XuZLL00025} &3.94	&3.72&11.22 &8.31	&20.32	&20.32 &26.21	&35.62	&38.91\\
 &\textbf{AAD (Ours)} &10.60 &\textbf{12.71} &\textbf{13.20} &\textbf{15.36} &\textbf{25.39} &\textbf{38.08} &\textbf{34.37} &\textbf{39.10} &\textbf{41.22} 
\\
\hline
\hline
	\end{tabular}}

	\label{tab:comparison_A}
\end{table*}

\begin{table*}[th]
	\centering
  	\caption{Comparison with SOTA methods on 1-shot, 5-shot, and 20-shot setting on biology dataset (Animal) and medical datasets (Polyp and Eyeballss). The numbers in \textbf{Bold} indicate the best performance.}
  \scalebox{1.0}{
	\begin{tabular}{ll|ccc|ccc|ccc}
\hline
\hline
Backbone &Method  &\multicolumn{3}{c|}{Animal} &\multicolumn{3}{c|}{Eyeballs} &\multicolumn{3}{c}{Polyp}\\
&&1-shot &5-shot &20-shot &1-shot &5-shot &20-shot &1-shot &5-shot &20-shot\\
\hline
\multirow{8}{*}{ResNet-50}
&PFENet{~\color{gray}
{[TPAMI2022]}}~\cite{TianZSYLJ22} &32.89 &33.43 &32.74&9.06 &9.06 &9.14 &13.48 &13.73 &13.81
\\
  &DCAMA{~\color{gray}{[ECCV2022]}}~\cite{ShiWZLNCMZ22}&48.80 &\textbf{56.53} &\textbf{59.06} &9.51&9.55 &9.60 &14.77 &14.01 &14.32\\
&HDMNet{~\color{gray}{[CVPR2023]}}~\cite{PengTWWLSJ23} &44.28 &48.95 &50.71 &9.04 &9.12 &9.11 &12.77 &13.84 &13.40
\\
&PFENet++{~\color{gray}{[TAPMI2024]}}~\cite{LuoTZYTJ24} &42.22 &44.29 &44.64 &9.17 &9.18 &9.15  &13.79 &13.47 &13.76\\
&HMNet{~\color{gray}{[NIPS2024]}}~\cite{XuLZL00024}&40.36	&49.03	&48.21	&8.98	&9.10	&8.98	&11.75	&13.61	&13.32\\
&ABCDFSS{~\color{gray}{[CVPR2024]}}~\cite{Herzog24}&25.78	&33.81	&34.82	&12.02	&11.78	&12.05	&15.34	&14.89	&14.29
\\
&NTRENet++{~\color{gray}{[TCSVT2025]}}~\cite{LiuLWCAYH25} &38.66	&43.22	&43.89	&4.21	&7.64	&10.88	&18.29	&18.49	&18.62
\\
 &\textbf{AAD (Ours)} &\textbf{52.85} &56.29 &58.79 
 &\textbf{13.04} &\textbf{13.71} &\textbf{13.85} &\textbf{21.93} &\textbf{23.91} &\textbf{23.89} \\
 \hline
 \multirow{3}{*}{ResNet-101} 
  &DCAMA{~\color{gray}{[ECCV2022]}}~\cite{ShiWZLNCMZ22}&55.51 &59.35 &59.06 &10.20&10.75 &11.21 &21.71 &25.13 &31.85\\
 &SCCAN{~\color{gray}{[ICCV2023]}}\cite{abs-2308-09294} &47.93&57.05  &57.61&9.01 &8.90 &9.12 &12.64 &13.47 &12.99\\
 &\textbf{AAD (Ours)}  &\textbf{61.11} &\textbf{63.71}  &\textbf{64.89} &\textbf{11.15} &\textbf{12.00} &12.33&\textbf{28.98}  &\textbf{37.57} &\textbf{42.45}\\
  \hline
   \multirow{8}{*}{SWin-T-base} 
&SAM{~\color{gray}{[arxiv2023]}}~\cite{abs-2304-02643}&16.98 &16.98 &19.01  &8.93 &8.93 &8.93 &15.83 &15.83 &15.83\\
  &DCAMA{~\color{gray}{[ECCV2022]}}~\cite{ShiWZLNCMZ22}&61.91 &64.88 &66.25   &9.86 &9.85 &9.87 &\textbf{33.28} &26.28 &31.70\\
  &Matcher{~\color{gray}{[ICLR2024]}}~\cite{LiuZL0WS24}&9.30&9.32 &-- &9.30	&9.52 &-- &15.00 &14.78 &--\\
  &MGF-SAM{~\color{gray}{[NIPS2024]}}~\cite{ZhangGJLW24}&9.44	&10.09	&11.40  &9.13	&9.22	&9.25 &14.96	&14.62	&14.37\\
 &GRPN{~\color{gray}{[AAAI2025]}}~\cite{peng2025sam}&38.66	&43.22	&43.89 &10.99	&11.18	&11.16 &18.29	&18.49	&18.62\\
 &FCP{~\color{gray}{[AAAI2025]}}~\cite{park2025foreground}&42.67	&45.81	&50.47	&8.96	&8.97	&8.97	&12.75	&12.56	&12.50\\
 &FS-SAM{~\color{gray}{[ICML2025]}}~\cite{XuZLL00025} &57.09	&59.06	&\textbf{69.54} &9.25	&9.42	&9.16 &17.76	&21.18	&21.72\\
 &\textbf{AAD (Ours)}  &\textbf{63.26} &\textbf{65.85} & 65.54  &\textbf{11.47} &\textbf{12.46} &\textbf{12.72} &32.85 &\textbf{51.79} &\textbf{59.68} \\  
\hline
\hline

	\end{tabular}}
	\label{tab:comparison_B}
\end{table*}

\begin{table*}[th]
	\centering
  	\caption{{\color{mycolor}Comparison with SOTA  methods on 1-shot, 5-shot, and 20-shot setting on Lunar terrain (Astronomy), and City satellite (Geography) datasets. The 'Mean’ refers to the average results across all eight datasets in the ER-FSS benchmark. The numbers in \textbf{Bold} indicate the best performance.}}
  \scalebox{1.0}{
	\begin{tabular}{ll|ccc|ccc|ccc}
\hline
\hline
Backbone &Method   &\multicolumn{3}{c|}{Lunar terrain}&\multicolumn{3}{c|}{City Satellite} &\multicolumn{3}{c}{Mean}\\
&&1-shot &5-shot &20-shot &1-shot &5-shot &20-shot &1-shot &5-shot &20-shot\\
\hline
\multirow{8}{*}{ResNet-50}
&PFENet{~\color{gray}
{[TPAMI2022]}}~\cite{TianZSYLJ22} &7.02 &7.65 &7.98 &4.91 &4.66 &4.50 &11.06 &11.13 &11.10\\
  &DCAMA{~\color{gray}{[ECCV2022]}}~\cite{ShiWZLNCMZ22} &5.28 &6.86 &8.47  &8.06 &6.58 &3.93 &14.60 &15.99 &17.09\\
&HDMNet{~\color{gray}{[CVPR2023]}}~\cite{PengTWWLSJ23}  &3.46 &5.16 &4.93 &5.16 &6.73 &9.21 &12.30 &13.66 &14.91\\
&PFENet++{~\color{gray}{[TAPMI2024]}}~\cite{LuoTZYTJ24} &5.74 &4.24 &7.89 &7.73 &10.35 &7.72  &11.36 &11.81&12.19\\
&HMNet{~\color{gray}{[NIPS2024]}}~\cite{XuLZL00024} &4.13	&5.50	&9.39	&8.42	&9.41	&10.17 &14.09&12.85&14.71\\
&ABCDFSS{~\color{gray}{[CVPR2024]}}~\cite{Herzog24} &6.87	&7.77	&8.49	&8.13	&9.43	&9.91 &11.50&12.94&14.79\\

&NTRENet++{~\color{gray}{[TCSVT2025]}}~\cite{LiuLWCAYH25} &6.96	&9.20	  &10.73	&0.97	&2.01	&1.98 &9.00&12.73&16.15\\
 &\textbf{AAD (Ours)} &\textbf{13.04} &\textbf{14.57} &\textbf{16.16} &\textbf{9.91} &\textbf{10.78} &\textbf{11.38} 
 &\textbf{19.24} &\textbf{21.90}  &\textbf{23.84}\\
 \hline
 \multirow{3}{*}{ResNet-101} 
  &DCAMA{~\color{gray}{[ECCV2022]}}~\cite{ShiWZLNCMZ22} &4.91 &7.62 &8.23 &9.48 &10.00 &11.25 &16.79 &20.03 &21.12\\
 &SCCAN{~\color{gray}{[ICCV2023]}}\cite{abs-2308-09294} &6.34 &7.53 &6.78 &6.17 &9.00  &5.17 &13.94 &16.83 &15.38\\
  &\textbf{AAD (Ours)}&\textbf{9.35} &\textbf{10.42} &\textbf{14.13} &\textbf{10.28}  &\textbf{10.91} &\textbf{11.43} 
   &\textbf{21.22} &\textbf{25.03}  &\textbf{27.21}\\
  \hline
   \multirow{8}{*}{SWin-T-base} 
&SAM{~\color{gray}{[arxiv2023]}}~\cite{abs-2304-02643} &4.45 &4.45 &4.45&8.68&8.68&8.68 &9.62 &9.62 &11.03\\
  &DCAMA{~\color{gray}{[ECCV2022]}}~\cite{ShiWZLNCMZ22}    &8.16 &12.85 &15.33 &9.65 &11.47 &12.20 &21.48 &25.10 &28.29\\
  &Matcher{~\color{gray}{[ICLR2024]}}~\cite{LiuZL0WS24}&7.76	&8.17&-- &3.66	&3.86&-- &8.93	&9.05 &-- \\
  &MGF-SAM{~\color{gray}{[NIPS2024]}}~\cite{ZhangGJLW24}&8.99	&10.32	&12.07 &4.41	&5.69	&6.47 &9.84	&11.29	&12.07\\
 &GRPN{~\color{gray}{[AAAI2025]}}~\cite{peng2025sam} &\textbf{13.92}	&16.59	&18.07 &7.66	&8.16	&8.04 &16.61 &19.19&21.29\\
 &FCP{~\color{gray}{[AAAI2025]}}~\cite{park2025foreground} &3.39	&3.37	&3.25	&8.86	&8.87	&8.85 &13.28 &12.92 &14.94\\
 &FS-SAM{~\color{gray}{[ICML2025]}}~\cite{XuZLL00025} &11.58	&15.16	&18.87 &9.20	&11.55	&\textbf{15.28}&17.92 &22.00&25.63\\
   &\textbf{AAD (Ours)} &9.86 &\textbf{17.13} &\textbf{21.21} &\textbf{10.00} &\textbf{12.51} &14.20 
&\textbf{23.47} &\textbf{29.62}  &\textbf{33.23}\\
   
\hline
\hline
	\end{tabular}}

	\label{tab:comparison_C}
\end{table*}

\subsection{Experiment Setup}
\noindent\textbf{Datasets.}
We utilize the general datasets PASCAL~\cite{ShabanBLEB17}, MSCOCO~\cite{NguyenT19}, and FSS-1000~\cite{LiWCTT20} with SBD augmentation as pre-training data, then evaluate the trained models on the proposed ER-FSS benchmark datasets, as proposed in Sec.\ref{sec:benchmark}. Note that for a fair comparison, we exclude classes that overlap between the pre-train datasets and the evaluation datasets.

\noindent\textbf{Training and Testing Strategy.} We evaluate two categories of methods using distinct protocols tailored to their architectural paradigms. For pre-trained segmentation models (e.g., SAM~\cite{abs-2304-02643}, Matcher~\cite{LiuZL0WS24}, MGF-SAM~\cite{ZhangGJLW24}, FS-SAM~\cite{XuZLL00025}), we follow a transfer learning strategy: the models undergo end-to-end pre-training on the base dataset, and during evaluation, they are fine-tuned on the support images before predicting query masks. For FSS-specific models (e.g., PFENet~\cite{TianZSYLJ22}, DCAMA~\cite{ShiWZLNCMZ22}, HDMNet~\cite{PengTWWLSJ23}, SCCAN~\cite{abs-2308-09294}, PFENet++~\cite{LuoTZYTJ24}, HMNet~\cite{XuLZL00024}, ABCDFSS~\cite{Herzog24}, NTRENet++~\cite{LiuLWCAYH25}, GRPN~\cite{peng2025sam}), we adopt the standard meta-learning paradigm for both training and testing. Evaluation follows the protocol in~\cite{LiWCTT20}: we report the average mean-IoU over 2 runs~\cite{VinyalsBLKW16} with different random seeds, where each run consists of 1,000 tasks per dataset to ensure statistical reliability.

\noindent\textbf{Evaluation Metric.}
We assess segmentation performance using the metric of Mean Intersection over Union (mIoU), a measure defined as the mean IoUs across all image classes. To compute the IoU for each category, we utilize the formula $IoU = \frac{TP}{TP + FP + FN}$, where $TP$, $FP$, and $FN$ represent the count of true positive, false positive, and false negative pixels in the predicted segmentation masks.

\noindent\textbf{Implementation Details.}
For a fair comparison, we employ Swin-transformer~\cite{LiuL00W0LG21}, ResNet-50~\cite{HeZRS16}, and ResNet-101~\cite{HeZRS16} as feature extraction networks, all of which are initialized with the weights pre-trained on ILSVRC~\cite{RussakovskyDSKS15} and kept frozen during the training process, following the previous works~\cite{LiWCTT20,BoudiafKZPAD21}. 
The input dimensions for support and query images are set at $384 \times 384$. ResNet-50 and ResNet-101 feature maps have channel dimensions of 256, 512, 1024, and 2048, while Swin-transformer feature maps have dimensions of 192, 384, 768, and 1536.
For ResNet-50-based AAD, the number of learnable initialization queries is set to 15, and the interaction with $\frac{1}{8}$, $\frac{1}{16}$, and $\frac{1}{32}$ dimensions of support and query features. In the case of the ResNet-101-based model, the number of queries is set to 20, with feature interactions at $\frac{1}{32}$ dimension Meanwhile, for the Swin-transformer-based model, 15 queries are used, with feature interactions at $\frac{1}{8}$, $\frac{1}{16}$, and $\frac{1}{32}$ dimensions.
The decoder is configured with $2$ convolutional layers, and between each module, bilinear interpolation is applied to upsample the feature maps by a factor of $2$, resulting in a total of $2$ upsampling functions. These networks were implemented using PyTorch, with AdamW~\cite{LoshchilovH19} as the optimizer, a learning rate of $1e-4$, and a weight decay of $0.05$.
During training, the batch size is set to $120$, and the training process ran on $8$ NVIDIA A800-SXM4-80GB GPUs in parallel, with subsequent evaluation on one GPU.

\subsection{Comparison with SOTA Models}

We present a comprehensive comparison of our proposed AAD framework against SOTA methods across eight diverse datasets in the ER-FSS benchmark. As summarized in Table~\ref{tab:comparison_C} (Mean), AAD consistently outperforms existing approaches across all backbone architectures and shot settings. Notably, with the ResNet-50 backbone, AAD achieves average mean-IoU scores of 19.24\%, 21.90\%, and 23.84\% for 1-shot, 5-shot, and 20-shot tasks, respectively. This represents a substantial improvement over the previous best-performing method, NTRENet++~\cite{LiuLWCAYH25} (e.g., +10.24\% in the 1-shot setting). Similar trends are observed with the ResNet-101 backbone, where AAD surpasses the strong baseline DCAMA~\cite{ShiWZLNCMZ22} by margins of 4.43\%, 5.00\%, and 6.09\% across the respective shot settings. These results validate the efficacy of our adaptive attention mechanism in learning generalized representations.

Furthermore, AAD demonstrates superior robustness in challenging environments characterized by high noise and domain shifts, such as industrial defects and medical imaging. As shown in Table~\ref{tab:comparison_A}, on the \textit{Steel defect} dataset---which features subtle targets and complex textures---AAD with ResNet-50 achieves a 1-shot IoU of 10.44\%, significantly outperforming NTRENet++ (4.41\%) and ABCDFSS (7.90\%). Similarly, in the medical domain (Table~\ref{tab:comparison_B}), AAD excels on the Polyp dataset, achieving a remarkable 21.93\% in the 1-shot setting compared to 18.29\% for NTRENet++. These gains highlight the model's ability to suppress environmental noise and accurately localize targets even when visual cues are obscure, effectively mitigating the domain gap between training and testing distributions.

Finally, we compare AAD against recent Transformer-based and SAM-based methods, which typically benefit from large-scale pre-training. Despite the strong transfer capabilities of models like SAM~\cite{abs-2304-02643} and FS-SAM~\cite{XuZLL00025}, AAD maintains a competitive edge. On the \textit{Animal} dataset (Table~\ref{tab:comparison_B}), AAD achieves a 1-shot performance of 63.26\%, surpassing FS-SAM (57.09\%) and GRPN (38.66\%). Even in the aggregate mean results (Table~\ref{tab:comparison_C}), our Transformer-based variant leads the leaderboard with 23.47\% (1-shot) and 29.62\% (5-shot), outperforming the sophisticated FS-SAM by 5.55\% and 7.62\%, respectively. This superiority suggests that while pre-training provides a strong foundation, our proposed attention distillation mechanism is crucial for adapting to specific, unseen tasks where intra-class variance and environmental interference are prevalent.

\begin{figure}[thb]
	\centering
	\includegraphics[width=1.0\linewidth]{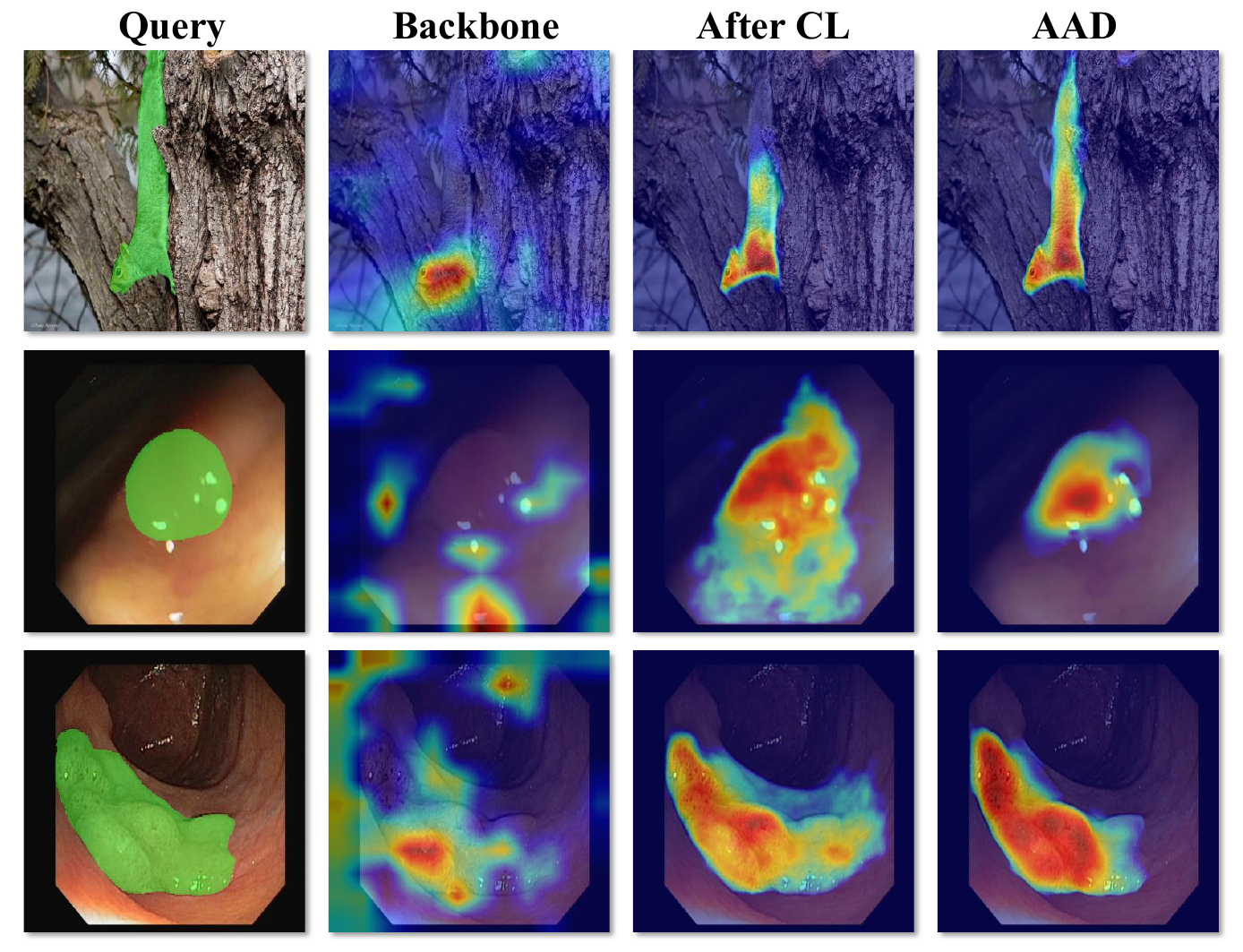}
	\caption{{\color{newcolor}Visualization of query maps at different stages under various environmental perturbations using the Swin-Transformer backbone. ``After CL" denotes the output following the correlation learning module.}}
	\label{Fig:feature}
\end{figure}

\begin{figure*}[thb]
	\centering
	\includegraphics[width=0.90\linewidth]{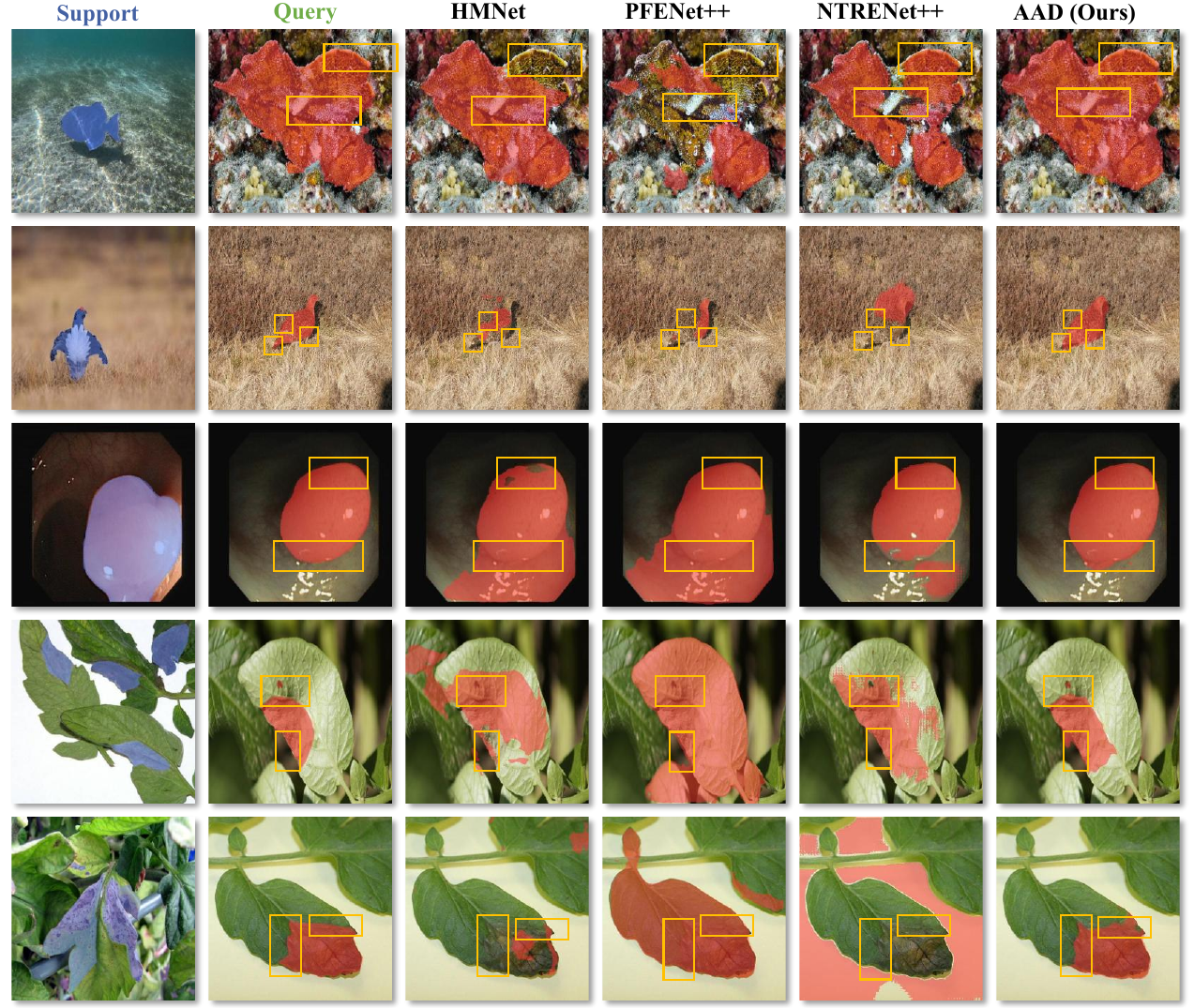}
	\caption{{\color{newcolor}Comparison of segmentation results between our method and SOTA methods on various evaluation datasets and under multiple difficult scenarios.}}
	\label{Fig:results}
\end{figure*}

\subsection{Further Analysis}

\noindent\textbf{(1) Ablation Study for Proposed Modules.} 
We present experimental results for the proposed elements, the correlation learner (CL) module and AAD. Tab.~\ref{tab:ablation_A} indicates that incorporating CL improves performance by 6\%, 17.8\%, 14.0\%, 18.2\%, 19.4\%, and 38.5\% on three datasets compared to the baseline. The additional inclusion of AAD results in further improvements of 7.6\%, 17.8\%, 14.0\%, 18.2\%, 19.4\%, and 38.5\% over the baseline. It can be proved that {\color{newcolor}AAD} enhances the results of both the baseline and baseline+CL, with more pronounced effects as the number of shots increases. This is attributed to AAD's ability to utilize more support images to generate more accurate class discriminative information.

\begin{table}[th]
	\centering
  	\caption{Ablation Study of correlation learning (CL) module and class discriminative information learner (AAD) with Swin-transformer. The baseline refers to HSNet~\cite{MinKC21}.}
  \scalebox{1.0}{
	\begin{tabular}{c|cc|cc|cc}
\hline
\hline
Method &\multicolumn{2}{c|}{Steel defect}&\multicolumn{2}{c|}{Leaf disease}&\multicolumn{2}{c}{Polyp}\\
&1-shot &5-shot&1-shot &5-shot&1-shot &5-shot\\
\hline
Baseline &7.72 &7.56 &20.31&20.94&13.48&13.13\\
+CL&14.29& 21.24 &27.73 &29.44 &22.32 &26.28\\
AAD&\textbf{15.36}&\textbf{25.39}&\textbf{34.37} &\textbf{39.10} &\textbf{32.85} &\textbf{51.79}\\
\hline
\hline
	\end{tabular}}

	\label{tab:ablation_A}
\end{table}

Moreover, Fig.~\ref{Fig:feature} visualizes the evolution of query feature maps at three critical stages: post-Backbone, post-Correlation Learner (CL), and post-AAD. These visualizations cover a spectrum of challenging environmental perturbations, including image blur, small and slender objects, occlusion, and camouflage. Observing the progression, it is evident that while the interactive learning within the CL module offers a preliminary filtration of irrelevant background noise compared to the raw backbone features, it often struggles to fully isolate the target in complex scenarios. In contrast, the integration of the AAD module proves decisive. By distilling robust class semantics, AAD significantly sharpens the model's focus, enabling it to accurately localize and highlight difficult targets even when they are obscured by camouflage or severe occlusion, thereby validating its superior environmental robustness.

\begin{table}[th]
	\centering
  	\caption{{\color{newcolor}Ablation study on the effectiveness and efficiency of the number of learnable queries.}}
  \scalebox{0.90}{
	\begin{tabular}{c|cc|cc|cc|c}
\hline
\hline
Queries &\multicolumn{2}{c|}{Steel defect}&\multicolumn{2}{c|}{Leaf disease}&\multicolumn{2}{c|}{Polyp}&GFLOPs\\
num(\#)&1-shot &5-shot&1-shot &5-shot&1-shot &5-shot&\\
\hline
0 &7.72 &7.56 &20.31&20.94&13.48&13.13&275\\
5 &14.88 &19.38 &33.53 &38.53 &36.35 &49.31 &275\\
15 &\textbf{15.36} &\textbf{25.39} &\textbf{34.37} &39.10 &32.85 &\textbf{51.79} &275\\
30 &14.96 &21.40&33.04 &37.49 &31.80 &43.39 &276\\
50 &13.83 &20.89&33.13 &35.64 &\textbf{36.51} &51.26 &276\\
100 &14.82 &14.23&32.36 &\textbf{42.55} &31.15 &32.99 &277\\
\hline
\hline
	\end{tabular}}

	\label{tab:ablation_B}
\end{table}

\noindent\textbf{(2) The Effectiveness and Efficiency of the Number of Learnable Queries.} 
{\color{newcolor}Table~\ref{tab:ablation_B} investigates the impact of the number of learnable queries on both performance and computational efficiency. Notably, the baseline without any learnable queries ($N=0$) yields exceptionally poor results across all datasets, underscoring their critical role in the framework. Introducing even a small number of queries ($N=5$) leads to a substantial performance leap. As $N$ increases to 15, the model achieves optimal or near-optimal results in most scenarios, securing the highest scores on the Steel defect (15.36\% for 1-shot, 25.39\% for 5-shot) and Leaf disease (34.37\% for 1-shot). Beyond $N=15$, the performance exhibits a general trend of slight decline or irregular fluctuation, suggesting that an excessive number of queries may introduce redundant information or noise. Consequently, we select $N=15$ as the optimal default setting. Importantly, increasing $N$ from 0 to 100 incurs negligible computational overhead, with GFLOPs only marginally increasing from 275 to 277. Since these queries function as low-dimensional vectors interacting within a lightweight module, their contribution to the total computational cost is minimal, ensuring that the model achieves robust performance with high efficiency.}

\begin{table}[th]
	\centering
  	\caption{Ablation study for different combinations methods of support features $f^{s}$, query features $f^{q}$, support masks $M^{s}$, and query aggregation masks $M^{q'}$.}
  \scalebox{0.95}{
	\begin{tabular}{l|cc|cc|cc}
\hline
\hline
Combination &\multicolumn{2}{c|}{Animal}&\multicolumn{2}{c|}{Steel defect}&\multicolumn{2}{c}{City satellite}\\
&1-shot &5-shot&1-shot &5-shot&1-shot &5-shot\\
\hline
Baseline &51.80 &55.92 &7.72 &7.56&9.61 &10.08 \\
Maskaad (Eq.\ref{eq:concat}) &61.35	&62.92	&14.38&24.93&9.41&11.59 \\
Concat (Eq.\ref{eq:add}) &61.47&63.52 &14.37&20.94&9.34&11.29\\
Ours (Eq.\ref{eq:foreground}--\ref{eq:transformer}) &\textbf{63.26}	&\textbf{65.85}	&\textbf{15.36}&\textbf{25.39}&\textbf{10.00}&\textbf{12.51}\\
\hline
\hline
	\end{tabular}}

	\label{tab:ablation_C}
\end{table}

\noindent\textbf{(3) Ablation Study for Feature Combination Methods.} 
In the class discriminative information learner, learnable queries interact with support features $f^{s}$, query features $f^{q}$, support mask $M^{s}$, and query aggregation masks $M^{q'}$ to learn category information. We employ a combined approach using Eq.~\ref{eq:foreground} and Eq.~\ref{eq:transformer}. Tab.~\ref{tab:ablation_C} presents two alternative feature fusion methods. Maskaad (Eq.~\ref{eq:add}) draws inspiration from Mask2former, where features are first attended to and then combined with the mask. Concat (Eq.~\ref{eq:concat}) involves concatenating features and masks before feeding them into ADG.The results in the table demonstrate the comprehensive superiority of our approach over the other two methods, indicating that directly identifying foreground features allows the model to learn more accurate category information.
\begin{equation}\label{eq:add}
{\color{newcolor}\tilde{q}^{m}_{n} = MLP[softmax(M^{q'}_{i}+\frac{\tilde{q}^{m}_{n-1}(f^{s}_{i}\otimes M^{s}_{i})}{\sqrt{d_{i}}}f^{q}_{i})+\tilde{q}^{m}_{n-1}]}
\end{equation}

\begin{equation}\label{eq:concat}
\begin{aligned}
{\color{newcolor}\tilde{q}^{m}_{n} =MLP\{softmax[(M^{s}_{i},M^{q'}_{i})+}\\
{\color{newcolor}\frac{\tilde{q}^{m}_{n-1}(f^{s}_{i},f^{q}_{i})}{\sqrt{d_{i}}}(f^{s}_{i},f^{q}_{i})]+\tilde{q}^{m}_{n-1}\}},
\end{aligned}
\end{equation}
Where $(M^{s}{i},M^{q'}{i})$ refer to the concatenation of two features.

\begin{table}[th]
	\centering
  	\caption{Comparison of experimental results with different backbones.}
  \scalebox{1.0}{
	\begin{tabular}{l|ccc|ccc}
\hline
\hline
Backbone  &\multicolumn{3}{c|}{Animal} &\multicolumn{3}{c}{Polyp}\\
&1-shot &5-shot &20-shot &1-shot &5-shot&20-shot\\
\hline
VGG-16&43.61&47.00&47.76&19.29&21.66&21.91\\
ResNet-50&52.85&56.29&58.79&21.93&23.91&23.89\\
ResNet-101&61.11&63.71&64.89&28.98&37.57&42.45\\
ViT-base&52.83&52.84 &53.67&20.72 &20.72&20.72 \\
SWin-T-base&\textbf{63.26} &\textbf{65.85} &\textbf{65.54} &\textbf{32.85} &\textbf{51.79} &\textbf{59.68}\\
\hline
\hline
	\end{tabular}}

	\label{tab:ablation_D}
\end{table}

\noindent\textbf{(4) Comparison Results for Multiple Backbones.}
Simultaneously, we showcase the results of our method on various backbones, all of which are loaded with pre-trained weights from ImageNet. ViT~\cite{DosovitskiyB0WZ21}, on the other hand, is initialized with pre-training parameters from CLIP~\cite{RadfordKHRGASAM21}. From Tab.~\ref{tab:ablation_D}, it can be observed that the results are optimal for ResNet-101 and Swin-Transformer, showing stability across multiple datasets and settings.

\begin{table}[ht]
	\centering
  	\caption{Comparison in comprehensive N-shot settings.}
  \scalebox{0.90}{
	\begin{tabular}{l|cccccc}
\hline
\hline
Method &\multicolumn{6}{c}{Polyp}\\
&1-shot &5-shot&10-shot &20-shot &30-shot &50-shot\\
\hline
FCN~\cite{LongSD15} &2.36& 2.98&3.30&4.17&7.99&14.50\\
SAM~\cite{abs-2304-02643} &15.83&15.83&15.83&15.83&15.83&15.83\\
DCAMA~\cite{ShiWZLNCMZ22} &22.32&26.28 &28.82&31.70&33.10&34.95\\
AAD&\textbf{32.85} &\textbf{51.79} &\textbf{56.64} &\textbf{59.68} &\textbf{61.09} &\textbf{62.25}\\
\hline
\hline
	\end{tabular}}
	\label{tab:ablation_E}
\end{table}

\noindent\textbf{(5) Comparison on More Settings of Shots.}
Also, we provide comparison results between our method and SOTA approaches in various shot settings, as shown in Tab.~\ref{tab:ablation_E}.
It demonstrates that AAD outperforms the second-ranked DCAMA by 10.6\%, 25.51\%, 27.82\%, 27.98\%, 27.99\%, and 27.30\% in 1-shot, 5-shot, 10-shot, 20-shot, 30-shot, and 50-shot settings, respectively. This confirms that AAD effectively utilizes the given support images, and the performance steadily improves as the shot number increases. In contrast, FCN and SAM show relatively minor performance improvements as the shot number increases, indicating that transfer learning-based pre-trained segmentation models struggle to transfer knowledge from simple samples to hard ones effectively.

\begin{table}[th]
	\centering
  	\caption{Comparison of Results and spend time of different K-shot inference methods on 20-shot with Swin-transformer.}
  \scalebox{1.10}{
	\begin{tabular}{l|cccc|c}
\hline
\hline
Method  &\multicolumn{1}{c}{Animal}&\multicolumn{1}{c}{Polyp
}&\multicolumn{1}{c}{Road}&\multicolumn{1}{c}{Leaf}&\multicolumn{1}{|c}{Time$\downarrow$(s)}\\
\hline
Vote    &64.81	&53.25 &\textbf{13.20} &\textbf{41.22}&2.89\\ 
Average &\textbf{65.54}  &\textbf{59.68}&12.14 &41.02&\textbf{2.51}\\
\hline
\hline
	\end{tabular}}
	\label{tab:ablation_F}
\end{table}

\noindent\textbf{(6) K-shot Inference.}
Additionally, when facing $K$-shot Inference, there are two commonly used methods: the voting method, which averages predictions for each support image and query separately, and the average method which averages the features of $K$ support images to generate a segmentation result for the query. Tab.~\ref{tab:ablation_F} displays the results of these two methods on AAD and compares their testing times in the 20-shot scenario. It indicates that the averaging method takes less time and outperforms the other by 6.4\% on the Polyp. The differences between the two testing methods are minimal for the other three datasets. In summary, the averaging method provides a higher cost-effectiveness.

\begin{table}[th]
	\centering
  	\caption{{\color{newcolor}Comparison of accuracy, GFLOPs, inference time per image and model parameters with Swin-Transformer between our model and the latest SOTA methods. }}
  \scalebox{1.0}{
	\begin{tabular}{l|cc|c|c|c}
\hline
\hline
Method &\multicolumn{2}{c|}{Mean}&GFLOPs &test\_t &Params\\
 &5-shot &20-shot&&(ms) &(MB)\\
\hline
 DCAMA~\cite{ShiWZLNCMZ22} &25.10 &28.29&260& 1.60 &239.67\\
NTRENet++~\cite{LiuLWCAYH25} &12.73&16.15&239&3.15 &269.05\\
 GRPN~\cite{peng2025sam}  &19.19&21.29&545&1.24 &206.01\\
\textbf{AAD (Ours)} 
 &\textbf{29.62}  &\textbf{33.23}&275 &1.65 &251.19\\
\hline
\hline
	\end{tabular}}

	\label{tab:para_com}
\end{table}

{\color{newcolor}\noindent\textbf{(7) Effectiveness Analysis.} As in Tab.~\ref{tab:para_com},  our AAD model demonstrates a highly optimal trade-off between performance and computational cost. Despite introducing specific interaction modules, AAD remains highly efficient with 275 GFLOPs, significantly lower than the Transformer-based GRPN (545 GFLOPs). Furthermore, its inference time (1.65 ms/image) and parameter count (251.19 MB) are strictly on par with existing methods like DCAMA and GRPN, indicating no severe computational bottlenecks. Most importantly, this efficiency is accompanied by substantial performance gains across all settings (e.g., 33.23 in 20-shot, far exceeding GRPN's 21.29). We have incorporated this comprehensive evaluation into the revised manuscript to highlight ADD's practical viability.}

\section{Conclusion}
{\color{newcolor} In this paper, we introduced the Environment-Robust Few-Shot Segmentation (ER-FSS) setting and benchmark to address the fragility of existing models under real-world perturbations. To tackle these challenges, we proposed the Adaptive Attention Distillation (AAD) framework, which distills robust class semantics via iterative support-query interactions. Despite achieving state-of-the-art robustness, we acknowledge certain limitations. In extremely low-data regimes like 1-shot scenarios, our performance occasionally trails behind specialized SOTA methods, likely because our learnable queries require more diverse support examples to effectively converge on a robust class representation. Future work will therefore focus on optimizing the model's efficiency and low-shot capability while significantly broadening the ER-FSS benchmark. We aim to incorporate a wider array of scenes and environmental perturbations, alongside other dimensions of difficulty—such as extreme intra-class diversity and acquisition variability—to construct a more holistic and challenging evaluation standard for practical segmentation systems.}

\bibliographystyle{IEEEtran}
\bibliography{Ref}

\begin{IEEEbiography}
[{\includegraphics[width=1in,height=1.25in,clip,keepaspectratio]{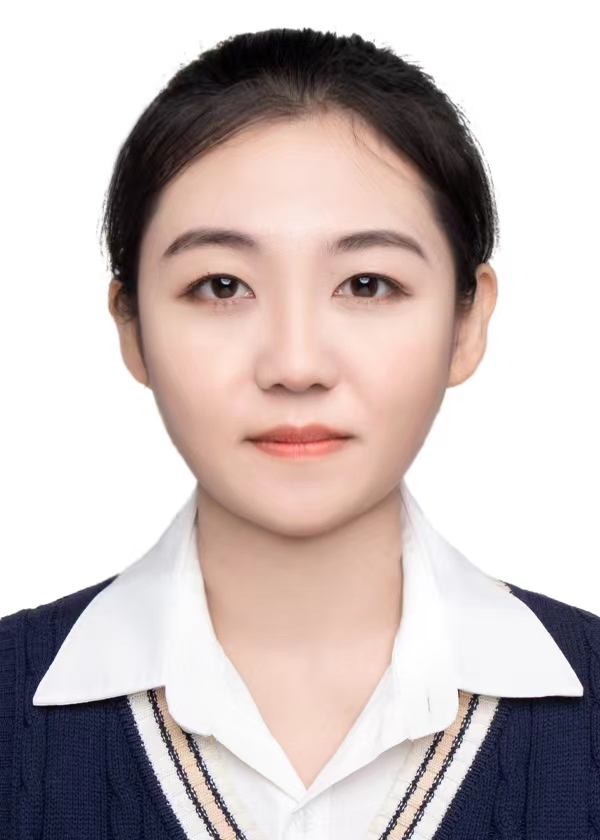}}] 
{\textbf{Qianyu Guo}} received her Ph.D. in Computer Science from Fudan University. She is currently an Assistant Professor at the Shanghai Institute of Virology, Shanghai Jiao Tong University. Her research interests include computer vision, AI for biology, and AI-driven drug discovery.
\end{IEEEbiography}

\begin{IEEEbiography}
[{\includegraphics[width=1in,height=1.25in,clip,keepaspectratio]{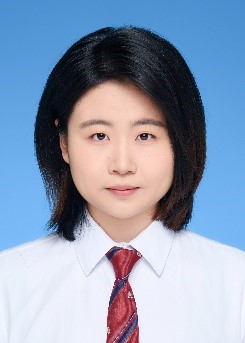}}] 
{\textbf{Jingrong Wu}} received her master's degree in Software Engineering from Southeast University. Her research interests include computer vision, model compression, and multimedia computing.
\end{IEEEbiography}

\begin{IEEEbiography}
[{\includegraphics[width=1in,height=1.25in,clip,keepaspectratio]{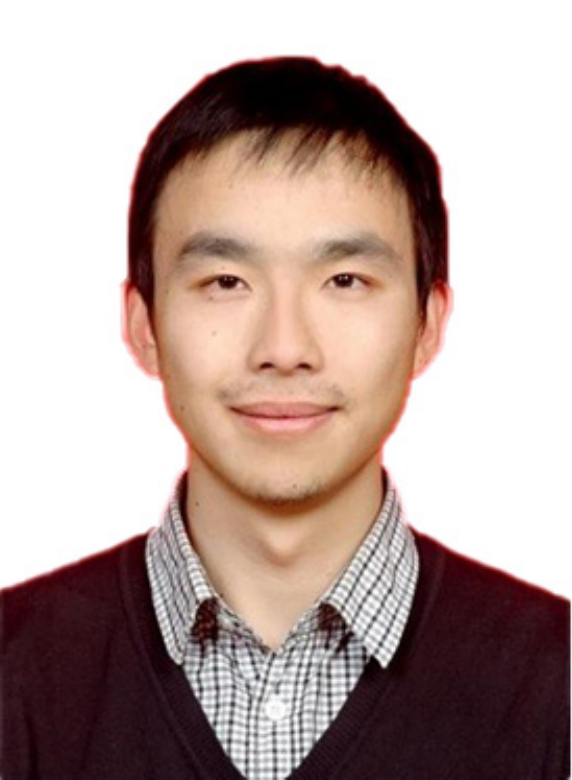}}] 
{\textbf{Jieji Ren}} received bachelor's and master's degrees in science from Harbin Institute of Technology, Harbin, China, in 2013 and 2015, respectively. He received a Ph.D. degree in mechatronic engineering from Shanghai Jiao Tong University, Shanghai, China, in 2022. Since November 2022, he has been with Shanghai Jiao Tong University as an assistant researcher. His research focuses on camera-based tactile sensing and its applications in soft robotics.
\end{IEEEbiography}

\begin{IEEEbiography}
[{\includegraphics[width=1in,height=1.25in,clip,keepaspectratio]{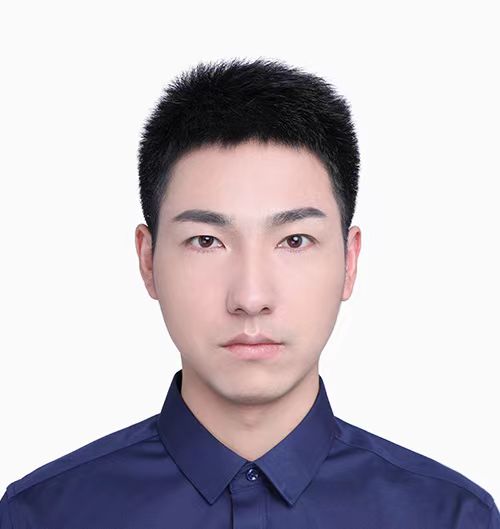}}] 
{\textbf{Weifeng Ge}} received the Ph.D. degree from The
University of Hong Kong in 2019. He is currently an Associate Professor with the School of Computer Science, at Fudan University. His current research interests include computer vision, deep learning, artificial general intelligence, and humanoid robots.
\end{IEEEbiography}

\begin{IEEEbiography}
[{\includegraphics[width=1in,height=1.25in,clip,keepaspectratio]{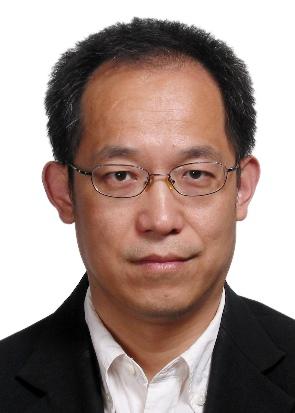}}] 
{\textbf{Wenqiang Zhang}} received a Ph.D. degree in
mechanical engineering from Shanghai Jiao Tong
University, China, in 2004. He is currently a Professor at the School of Computer Science, at Fudan University. His current research interests include computer vision and robot intelligence.
\end{IEEEbiography}

\vfill

\end{document}